\documentclass[letterpaper, 10 pt, conference]{ieeeconf}  

\IEEEoverridecommandlockouts                              

\usepackage{graphics} 
\usepackage{epsfig} 
\usepackage{mathptmx} 
\usepackage{times} 

\usepackage{amsmath}
\usepackage{amssymb}
\usepackage{graphicx}
\usepackage[breaklinks=true,bookmarks=false]{hyperref}
\title{\LARGE \bf
MonoNext: A 3D Monocular Object Detection with ConvNext
}

\author{Marcelo Eduardo Pederiva$^{1}$, José Mario De Martino$^{1}$ and Alessandro Zimmer$^{2}$
\thanks{This work was supported by The National Council for Scientific and Technological Development}
\thanks{$^{1}$Marcelo Eduardo Pederiva and José Mario De Martino are with School of Electrical and Computer Engineering,
        University of Campinas, Brazil
        {\tt\small marcelopederiva@gmail.com, martino@unicamp.br}}%
\thanks{$^{2}$Alessandro Zimmer is with the Center of Automotive Research on Integrated Safety Systems and Measurement Area, Technische Hochschule Ingolstadt,
        Germany
        {\tt\small Alessandro.Zimmer@thi.de}}%
}

\begin{document}

\maketitle
\thispagestyle{empty}
\pagestyle{empty}

\begin{abstract}

Autonomous driving perception tasks rely heavily on cameras as the primary sensor for Object Detection, Semantic Segmentation, Instance Segmentation, and Object Tracking. However, RGB images captured by cameras lack depth information, which poses a significant challenge in 3D detection tasks. To supplement this missing data, mapping sensors such as LIDAR and RADAR are used for accurate 3D Object Detection. Despite their significant accuracy, the multi-sensor models are expensive and require a high computational demand. In contrast, Monocular 3D Object Detection models are becoming increasingly popular, offering a faster, cheaper, and easier-to-implement solution for 3D detections. This paper introduces a different Multi-Tasking Learning approach called MonoNext that utilizes a spatial grid to map objects in the scene. MonoNext employs a straightforward approach based on the ConvNext network and requires only 3D bounding box annotated data. In our experiments with the KITTI dataset, MonoNext achieved high precision and competitive performance comparable with state-of-the-art approaches. Furthermore, by adding more training data, MonoNext surpassed itself and achieved higher accuracies. The code will be released at \url{https://github.com/marcelopederiva/MonoNext}.
\end{abstract}

\section{Introduction}

\begin{figure}[thpb]
    \centering
    \includegraphics[width=0.7\linewidth]{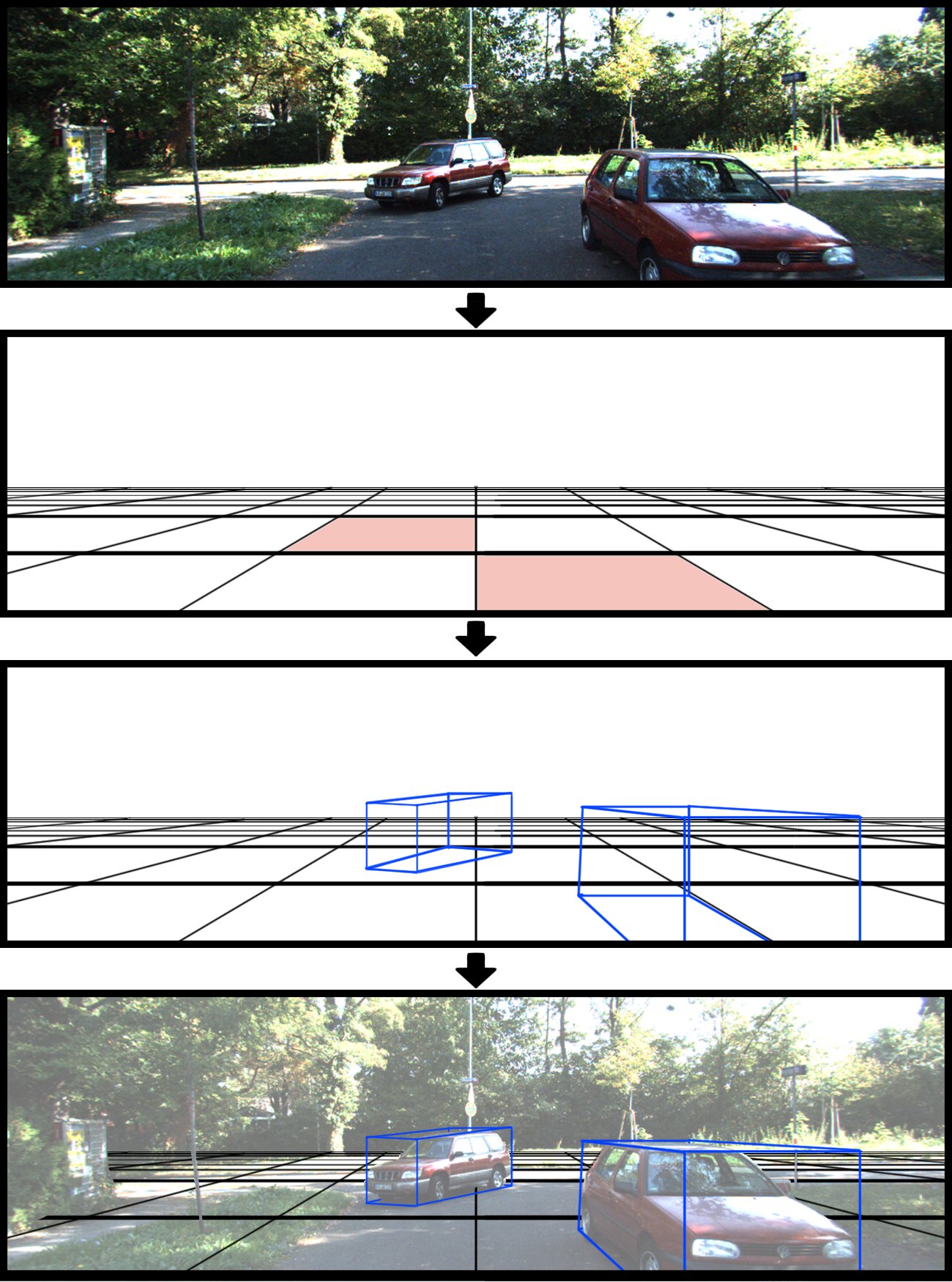}
    \caption{MonoNext Approach Overview. The model constructs a two-dimensional grid covering the area observed by the camera. Each cell of this grid will estimate if there is an object in it and its characteristics. Finally, the cells that have the highest confidence value of having an object are chosen.}
    \label{fig:approach}
\end{figure}

Object detection is a fundamental task in the field of Computer Vision. The ability to recognize and locate objects in both two and three dimensions plays a critical role in various areas, including robotic vision, safety applications, and autonomous driving.

Object detection based on images has experienced significant advancements over the years, attracting the attention of research centers, industries, and universities. While 2D object detection achieves high accuracy in real time, accurately estimating the three-dimensional characteristics of objects without deep information remains a challenging task.

Currently, state-of-the-art 3D Object Detection methods depend on the use of multiple sensors to estimate the characteristics of objects in a scene. By combining data from sensors that map the environment and cameras, these methods can achieve up to 80\% accuracy in detecting objects, as demonstrated by the KITTI benchmark \cite{KITTI}. However, sensors that provide this environment mapping, such as LIDARs, are expensive and computationally demanding. As a result, there is a growing interest in developing 3D detection models that rely solely on cameras in order to reduce cost and computational complexity.

While some existing works have explored the use of two cameras, from different perspectives, to improve 3D estimation (Stereo Object Detection) \cite{Liu_2021_YoloStereo3D,Chen_2022_DSGN}, this paper focuses on developing a 3D prediction model based on a single input image (Monocular Object Detection) to make the approach more accessible and expand its applications. Our model builds a Bird-Eye-View grid, where each cell grid contains the estimation of confidence, class, 3D position, 3D dimension, and yaw-angle of the object in it (Figure \ref{fig:approach}). The proposed approach is divided into three steps, including a light CNN backbone, followed by a ConvNext-based architecture, and a Multi-Task learning design for the object's characteristics estimation.  

The model was tested on the KITTI dataset to detect cars in urban streets, and its performance was comparable with current methods, achieving competitive accuracies in the validation set. Although the paper primarily presents a vehicle detection application for use in the field of Self-Driving cars, the model is highly flexible, allowing its implementation to detect different objects for a variety of applications.

The paper is organized as follows. Initially, a section of Related Works (Section \ref{related work}) reviews the state-of-the-art 2D and 3D detection models. Next, in section \ref{approach}, the main approach of our work is presented. This section shows a definition of the target problem and an explanation of our method, such as architecture, loss equation, and training parameters. In section \ref{experiment}, the results of our model on the KITTI dataset is presented along with a comparison of the results given by the state-of-the-art. Finally, in section \ref{conclusion}, a summary and the model's performance is discussed.

\section{Related Work}
\label{related work}
\subsection{2D Object Detection }

2D Object Detection is a Computer Vision task that involves locating and classifying objects within an image. There are two types of methods used for this task: two-stage and single-stage detectors. Two-stage detectors use a first step to find proposed regions and a second step to identify and classify objects in these regions, providing high accuracy but slower performance. Single-stage detectors use a straightforward approach for identifying objects, providing faster detections and real-time predictions.

The Region-based Convolutional Neural Network (R-CNN) pioneered the two-stage object detection approach by utilizing CNNs to improve object detection accuracy \cite{Girshick_2013_RCNN}. R-CNN uses a CNN, such as AlexNet \cite{AlexNet}, to propose regions of interest, which are then passed by another network to predict objects. The model featured the best accuracy at the time and inspired the development of derived versions. Fast R-CNN and Faster R-CNN are variations of R-CNN that further improved object detection performance by introducing better region proposal methods and approaches to detect faster \cite{Girshick_2015_FastRCNN, Ren_2015_FasterRCNN}. Furthermore, different two-stage methods emerged, presenting competitive performances. SPP-Net \cite{He_2014_SPPNet} proposes a Spatial Pyramid Pooling layer to handle variable input sizes and aspect ratios, reducing computational demands, and achieving comparable accuracy to R-CNN with faster detection times. The Feature Pyramid Network (FPN) concatenates feature maps from different scales to enhancing the detection of small objects and maintain accuracy at different scales \cite{Lin_2017_FPN}.

One-stage models have become increasingly popular due to their fast response time and straightforward approach. One of the most influential one-stage models is the Single Shot Multibox Detector (SSD) \cite{Liu_2016_SSD}. The SSD uses a cascading approach to detect objects based on different layers' dimensions in the Convolutional Network, achieving the same performance as two-stage models while maintaining real-time behavior.

Another widely known one-stage model is the You Only Look Once (YOLO) model \cite{Redmon_2015_YOLO}. The YOLO model uses a unique and straightforward architecture that predicts a classification grid based on the input image. Although the model initially did not have the best accuracy, it was very light and reached high detection speeds. Over the years, new versions of YOLO have emerged with changes in architecture and training to achieve high accuracy and low response times \cite{Redmon_2016_YOLOv2, Redmon_2018_YOLOv3, Bochkovskiy_2020_YOLOv4}.

The Swin-Transformer model is another approach to object detection that uses the Vision Transformer \cite{Liu_2021_SwinTranf}. The Transformer architecture has had a significant impact on Natural Language Processing (NLP) \cite{Vaswani_2017_Transformer}, and its vision-based approach achieves high accuracy in image classification. Although the model is still in its early stages, it has already shown competitive accuracy compared to other approaches.

\subsection{Monocular 3D Object Detection}

Unlike 2D detections, 3D object detection models are tasked with recognizing objects in a three-dimensional space, which requires accurate 3D localization, dimension, and angle rotation of the object. However, predicting these characteristics using only one image (monocular detection) and without depth information is challenging.

Several works propose different approaches for monocular detection \cite{Zhou_2019_CenterNet, Liu_2019_FQNet, Simonelli_2019_MonoDIS, Qin_2018_MonoGRNet, Liu_2020_SMOKE, Li_2020_RTM3D, Simonelli_2020_MoVi3D}. Some methods use the object’s position information in the image to detect 2D objects and then refine 3D estimates on it \cite{Chabot_2017_DeepManta}. The GS3D utilizes 2D detection models to project a 3D guidance and then refines the projection for the final detection of objects \cite{Li_2019_GS3D}. In contrast, MonoCon uses 2D detection data but proposes a direct approach to 3D estimation, where 2D information is only used to converge the loss error \cite{Liu_2022_MonoCon}. Mono3D uses 2D data along with class semantic and instance semantic information to refine the model \cite{Chen_2016_Mono3D}.

Alternatively, some approaches aim to detect objects in a single step \cite{Zhou_2021_MonoEF, Huang_2022_MonoDTR}. For example, M3D-RPN presents a parallel network approach for simultaneous prediction of 2D and 3D detections \cite{Brazil_2019_M3D-RPN}. The model uses deep-aware convolution techniques to learn spatial information. On the other hand, the Orthographic Feature Transform predicts a feature grid from the output of a front-end ResNet-18 to map the objects in the scene in a voxel-based representation. This grid projection is then refined by another network that predicts the object's characteristics in each grid \cite{Roddick_2018_Ortho}.

Other methods rely on geometric uncertainty projection approaches to estimate the depth distance of objects. Models like GUPNet \cite{Lu_2021_GUPNet}, MonoFlex \cite{Zhang_2021_MonoFlex} and MonoPair \cite{Chen_2020_MonoPair}, which utilize heatmap data, 3D and 2D labels to converge the training step, have shown impressive results. By combining various sources of information, the performance achieved by these models currently represents the state-of-the-art in Monocular 3D Object Detection.

\begin{figure*}[thpb]
\centerline{\includegraphics[width=\linewidth]{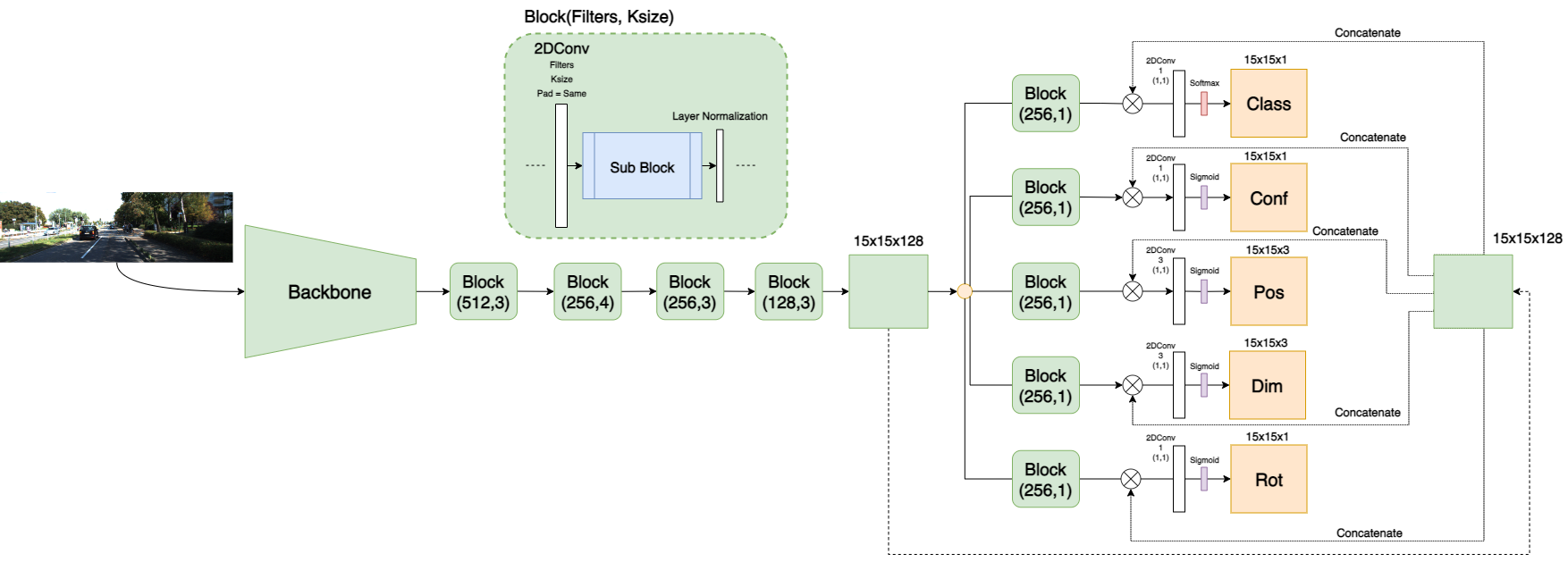}}
\caption{MonoNext Framework. The input image passes through a Feature Extractor, with a backbone and a ConvNext architecture. Then it branches into five tasks to predict each characteristic of the object in a 15$\times$ 15 grid BEV.}
\label{fig:framework}
\end{figure*}

\section{Our Approach}
\label{approach}

In this section, we introduce our proposed method for Monocular 3D Object Detection. Our approach utilizes Multi-Task Learning to estimate various characteristics of objects present in the scene based on a single input image, including 3D position (\textit{x,y,z}), 3D dimension (\textit{w,h,l}), class, and yaw-angle (\textit{$\theta$}).

Our goal is to create a two-dimensional grid to map the objects presented in the image, where each grid cell will provide confidence in having an object and its corresponding characteristics. As the target object (vehicles) has limited variations in the vertical axis and to ensure a lightweight model, we adopt a $15\times15$ grid in Bird-Eye-View to cover all the vehicles in the image. The prediction process of this grid will be shown below. 

First, an overview of the approach is presented to illustrate all processes of the model detection. Next, it is discussed the Feature Extractor, which is followed by the Multi-Task Learning step. Finally, the Loss Equation is presented.

\subsection{Framework Overview}

As shown in Figure \ref{fig:framework}, our approach is composed of 2 steps, a Feature Extractor and a Multi-Task Learning step. In the first stage, the model receives an RGB image as input and passes through 2 different CNNs, the MobileNetV2 as backbone, and a ConvNext-based network. The output of this stage already presents the Bird-Eye-View grid ($15\times15$) with 128 filters. Then, this layer is divided into 5 branches, where each branch will refine this output to predict a particular task: Class, Confidence, Position (x, y, z), Dimension (w, h, l), and Yaw-angle ($\theta$).

\subsection{Feature Extractor}

To extract features, our proposed method incorporates a known backbone along with a sequence of Convolutional Layers based on ConvNext.

Seeking to design a lightweight detection model, we opted for MobileNetV2 \cite{Sandler_2018_MobileNetv2} as the backbone of our Feature Extractor step. MobileNetV2 has demonstrated excellent performance while being one of the lightest feature extractors. This made it one of the best candidates to serve as the backbone of our model. However, having just the MobileNetV2 is insufficient for performing well in 3D detections. Therefore, our approach utilizes a ConvNext-based light architecture to complement this stage.

ConvNext \cite{Liu_2022_ConNext} has introduced a novel design of convolutional layers that have surpassed the accuracy of Transformers-based models. Despite its performance, the default ConvNext network requires high memory allocation. To address this limitation, we propose a simple ConvNext-based architecture for this stage.

\begin{figure}[thpb]
\centerline{\includegraphics[width=0.7\linewidth]{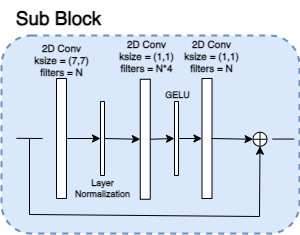}}
\caption{Sub Block architecture based on ConvNext. The block has 3 Convolutional Layers with a shortcut adding the output of the Convolutional sequence.}
\label{fig:subblock}
\end{figure}

Our ConvNext architecture is based on blocks, where each block is composed of a 2D Convolutional Layer (ConvLayer), a sub-block, and a Layer Normalization. Figure \ref{fig:subblock} illustrates the Sub-Block, which consists of a ConvLayer with kernel size 7 and N filters, followed by a Layer Normalization, a ConvLayer with kernel size 1 and 4*N filters, GELU activation function, another ConvLayer with kernel size 1 and N filters, and finally, this output is added to the input to create a shortcut.

The output of the Backbone passes by a sequence of 4 blocks, with each block having a specific number of filters and kernel size, represented as (512,3), (256,4), (256,3), and (128,3). 

\subsection{Multi-task Learning}

3D detection requires the prediction of many objects' characteristics. Thus, we apply a Multi-task Learning approach to estimate all required information simultaneously. The output of the Feature Extractor step is split into five tasks, and each branch passes through a residual ConvNext block(256,1). Subsequently, each task goes through a last Convolutional Layer, which utilizes an activation function, \textit{softmax} for Classification and \textit{sigmoid} for the other tasks.

In the end, we obtain a 15x15 horizontal-depth grid, where each grid cell contains nine values. One value represents the confidence score, and eight values estimate the 3D characteristics of the object. This approach allows us to map the objects in the image into the two-dimensional grid while simultaneously estimating the objects' characteristics.

\subsection{Loss Equation}

In the Multi-Task Learning step, our model generates multiple predictions simultaneously. To ensure the model convergence, we consider the errors of all tasks and sum them up in the total loss equation (Equation \ref{eq:losstotal}). As a consequence, our total loss will be the result of the sum of three specific losses: Confidence loss ($loss_{conf}$), Class loss ($loss_{class}$), Box loss ($loss_{box}$). In each equation, the variables with hat represent the ground truth and, without hat, the model's predictions. Additionally, to refine the relevance of each loss in the total loss value, a weight parameter ($\lambda$) was implemented.

We will consider only the cell grid responsible for detecting one object. Therefore, the $\mathsf{1}_i^{obj}$ parameter results in 1 if the cell is responsible for detecting an object. Otherwise, $\mathsf{1}_i^{obj}$ results in 0. The $\mathsf{1}_i^{noobj}$ represents the opposite of $\mathsf{1}_i^{obj}$.

\begin{equation}
	loss = loss_{conf} + loss_{class} + loss_{box}
	\label{eq:losstotal}
\end{equation}

To evaluate the Confidence score in each prediction, we used the L2 loss function. Equation \ref{eq:confidenceloss}) considers the confidence prediction at correct cells and the missed ones.
\begin{equation}
    \begin{split}
	loss_{conf} = & \lambda_{obj} \sum_{i=0}^{S_X*S_Z}\mathsf{1}_i^{obj} (\hat{C_i} - C_i)^2+\\ 
	& \lambda_{noobj} \sum_{i=0}^{S_X*S_Z}\mathsf{1}_i^{noobj} (\hat{C_i} - C_i)^2
	\label{eq:confidenceloss}
    \end{split}
\end{equation}

The Classification loss (Equation \ref{eq:classloss}) is represented by the square error of the prediction and ground truth.

\begin{equation}
	loss_{class} = \lambda_{class} \sum_{i=0}^{S_X*S_Z}\mathsf{1}_i^{obj}\sum_{c \in classes}(class_i - \hat{class}_i)^2
	\label{eq:classloss}
\end{equation}

Finally, the Box Loss is defined by the bounding box characteristics and the rotation angle. To evaluate the objects' dimensions and positions, we consider the square error of the IoU score of the prediction and ground truth and the square error of the rotation estimation.

\begin{equation}
	loss_{box} = \lambda_{IoU} \sum_{i=0}^{S_X*S_Z}\mathsf{1}_i^{obj}IoU +
	\lambda_{\theta} \sum_{i=0}^{S_X*S_Z}\mathsf{1}_i^{obj}(\theta_i - \hat{\theta}_i)^2
	\label{eq:boxloss}
\end{equation}

The weight values were optimized at the training step, avoiding overfitting in all tasks. As a result, the weights are defined as follows: $\lambda_{obj} = 5$; $\lambda_{noobj} = 1$; $\lambda_{class} = 1$;\\ $\lambda_{IoU} = 10$; $\lambda_{\theta} = 1$.

The $\lambda_{obj}$ requires a higher weight than $\lambda_{noobj}$ to incentive the training step to start detecting all existing objects in the scene. With low weight in $\lambda_{noobj}$, the learning process slowly disregards the wrong confidences. On the other hand, keeping a low weight in $\lambda_{obj}$, the training step did not converge to detect all objects.

\section{Experiments}
\label{experiment}

    The proposed method is evaluated on KITTI 3D Object Detection benchmark \cite{KITTI}. The dataset includes an open-access 7481 images for training and a 7481 closed test set. However, the test set submissions are limited. Consequently, the \cite{Chen_2017_CVPR} proposed a model's evaluation with the training set, splitting the open-access set into approximately 50\% for training (3712 images) and the rest for validation (3769 images).

    The results are evaluated on three different difficulty levels: Easy, Moderate, and Hard, representing a fully visible, part-occluded, and mostly-occluded object, respectively. In addition, the dataset presents multiple classes, such as Car, Truck, Van, Pedestrian, and Cyclist.
    
\subsection{Implementation and Training Parameters}

    The input image is resized to $480\times480$. The increase in contrast and horizontal flip was adopted as the data augmentation. Our model was trained in a GPU: RTX2080ti and CPU: intel i7 9700KF with batch size 8, achieving the best results with around 200 epochs. The AdamW optimizer was used with an initial learning rate of 1e-4 and weight decay of 1e-6. For performance comparison, the model was trained to recognize only one class (vehicles) around [-55/55m, 0/15m, 0/85m]. However, our model is flexible to detect multiple classes and in different environments.
    
\subsection{Main Results}
    
    As shown in Table I, we trained our model with different configurations of training sets: 50\% dataset (3712 images), 50\% dataset with data augmentation, 80\% dataset (5984 images), 80\% dataset with data augmentation. Furthermore, our model was not pre-trained with different 3D detection datasets. The performance was acquired training only with the provided data from the KITTI training set. As a consequence, our method performance is highly dependent on the amount of training data. 

    \begin{table}[thpb]
    \centering
    \label{tab:ablation}
    \begin{tabular}{l|c|c|c|c}
     & Easy           & Moderate       & Hard           & Average        \\ \hline
    MonoNext (50\%) & 13.49          & 5.61           & 7.79           & 8.96           \\
    MonoNext (50\% + data aug)      & 20.76          & 9.79           & 8.70           & 13.08          \\
    MonoNext (80\%) & 24.12          & 14.26          & 14.46          & 17.61          \\
    \hline
    \textbf{MonoNext (80\% + data aug)}    & \textbf{27.28} & \textbf{16.01} & \textbf{15.01} & \textbf{19.43} \\
    \hline
    \end{tabular}
    \caption{MonoNext performance with the increment of data.}    
\end{table}
    
    In the KITTI dataset, there are more Easy detections shown in the images. In this case, using 50\% of the dataset for training contributes to achieving great performance in this category and does not generalize the model enough for other detections (Moderate, Hard). Conversely, with 80\% of the dataset + data augmentation, our model achieves high performance in 3D detection based on monocular models. The model presents a proper generalization of each detection category and a higher performance in hard detections. The table shows that the model still has the performance to achieve better results with more training data.

    Also, in Table \ref{tab:perform}, you can find the model's overall performance with different metrics. Average Recognition represents vehicles recognized in the scene with an IoU\textgreater{}0.1. Considering correct detections with IoU\textgreater{}0.5, the model correctly detects 43\% of full visible vehicles. Furthermore, evaluating the recognition of objects, MonoNext reaches an accuracy of ~72\% in its detections.

    \begin{table}[thpb]
        \centering
        \begin{tabular}{l|c|c|c}
                                       & Easy   & Moderate & Hard   \\ \hline
        Mean IoU                       & 0.4624 & 0.2257   & 0.1361 \\
        mAP(IoU\textgreater{}0.7) (\%) & 27.28  & 16.01    & 15.01  \\
        mAP(IoU\textgreater{}0.5) (\%) & 43.21  & 29.26    & 26.5   \\
        Average Recognition (\%)       & 72.26  & 65.07    & 60.52 
        \end{tabular}
        \caption{MonoNext Overall Performance in Car detection.}
        \label{tab:perform}
    \end{table}

\begin{figure*}[thpb]
    \centerline{\includegraphics[width=\linewidth]{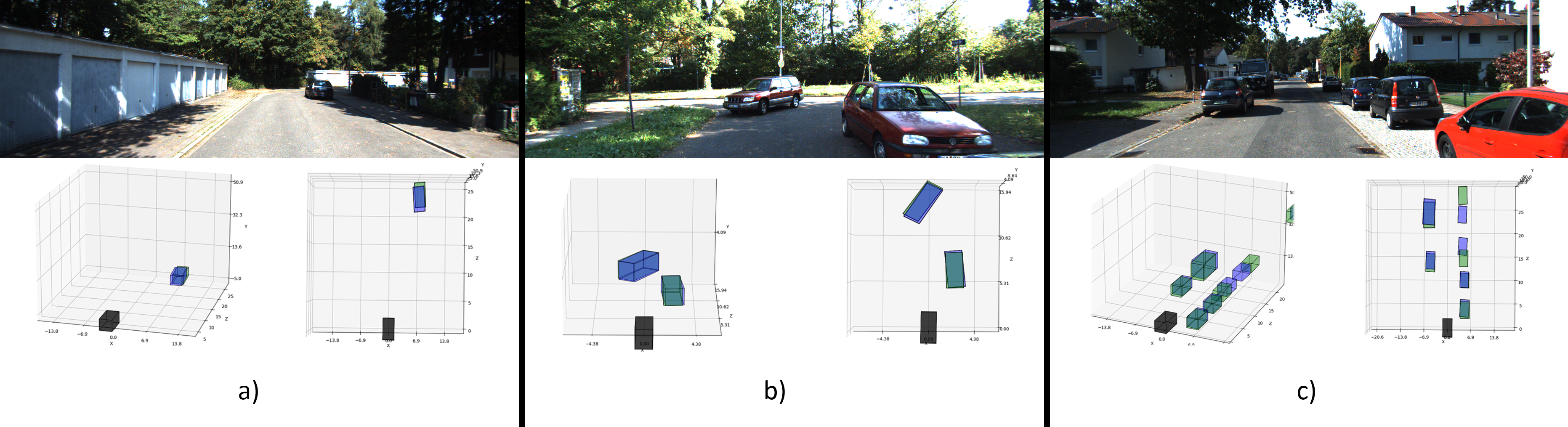}}
        \caption{The visualized result of MonoNext detections on the validation set. In Black is the Self-driving car and camera position, in Green are the Ground truth object boxes, and the Blue represents the MonoNext predictions. In each image: Above shows the input image; on the left side, there is a three-dimensional view; on the right, a Bird-Eye-View (BEV) of the detections.}
    \label{fig:all_img}
    \end{figure*}

    Figure \ref{fig:all_img} shows the output of MonoNext, demonstrating great and bad predictions to show the efficiency of our model. In Figures \ref{fig:all_img}a) and \ref{fig:all_img}c) we see that the model recognizes all vehicles in the scene, accurately predicts fully visible vehicles, and has difficulty estimating depth in cases of vehicles that present low brightness. On the other hand, in Figure \ref{fig:all_img}b) we observe that the model presents excellent precision in the detection of nearby vehicles, accurately predicting both the 3D bounding box and the yaw-angle of the vehicles.

\textbf{Comparison on the car category on validation set.}

In Table 3, we compare our model against the state-of-art in the KITTI validation set. Despite using a small dataset of 3712 images (50\%), MonoNext outperforms several images-only methods. We achieved competitive performance in Easy detections, with a 4.48\% increase in accuracy compared to MonoPair. While our model did not reach the same performance as GUPNet in all categories, it's important to note that GUPNet uses pre-trained weights, which helps in achieving higher performance even with a small dataset.

One of the advantages of MonoNext is that it does not require 2D labels for model convergence, unlike GUPNet. With only 3D feature annotations, MonoNext excels in its performance and proves to be a competitive option for monocular detections. Moreover, our method is simple and easy to implement.

\begin{table}[thpb]
\centering
\small
\label{tab:val}
\begin{tabular}{l|ccc|c}
\multicolumn{1}{c|}{}                 & \multicolumn{4}{c}{$AP_{3D}(IoU = 0.7)(\%)$}                                                                         \\ \hline
\multicolumn{1}{c|}{Models}            & \multicolumn{1}{c|}{Easy} & \multicolumn{1}{c|}{Moderate} & Hard  & Average                     \\ \hline
CenterNet \cite{Zhou_2019_CenterNet}                            & \multicolumn{1}{c|}{00.60}           & \multicolumn{1}{c|}{00.66}              & 00.77          & 00.68                     \\
FQNet \cite{Liu_2019_FQNet}                                 & \multicolumn{1}{c|}{02.77}          & \multicolumn{1}{c|}{01.51}              & 01.01          & 01.76                       \\
MonoDIS \cite{Simonelli_2019_MonoDIS}                               & \multicolumn{1}{c|}{11.06}         & \multicolumn{1}{c|}{07.60}               & 06.37          & 08.34                       \\
MonoGRNet \cite{Qin_2018_MonoGRNet}                             & \multicolumn{1}{c|}{11.90}          & \multicolumn{1}{c|}{07.56}              & 05.76          & 08.41                     \\
SMOKE \cite{Liu_2020_SMOKE}                                & \multicolumn{1}{c|}{14.03}         & \multicolumn{1}{c|}{09.76}              & 07.84          & 10.54                      \\
RTM3D \cite{Li_2020_RTM3D}                                & \multicolumn{1}{c|}{14.41}         & \multicolumn{1}{c|}{10.34}             & 08.77          & 11.17                      \\
M3D-RPN \cite{Brazil_2019_M3D-RPN}                              & \multicolumn{1}{c|}{14.53}         & \multicolumn{1}{c|}{11.07}             & 08.65          & 11.42                      \\
MoVi3D \cite{Simonelli_2020_MoVi3D}                               & \multicolumn{1}{c|}{14.28}         & \multicolumn{1}{c|}{11.13}             & 09.68          & 11.70                      \\
MonoPair \cite{Chen_2020_MonoPair}                             & \multicolumn{1}{c|}{16.28}         & \multicolumn{1}{c|}{12.30}              & 10.42         & 13.00                      \\ \hline
\multicolumn{1}{|l|}{\textbf{MonoNext (Ours)}} & \multicolumn{1}{c|}{20.76}         & \multicolumn{1}{c|}{09.79}              & 08.70          & \multicolumn{1}{c|}{13.08} \\ \hline
GupNet \cite{Lu_2021_GUPNet}                               & \multicolumn{1}{c|}{\textbf{22.76}}         & \multicolumn{1}{c|}{\textbf{16.46}}             & \textbf{13.72}         & \textbf{17.65}                     
\end{tabular}
\caption{Performance of the Car category on the KITTI validation set.}
\end{table}

\section{Conclusions}
\label{conclusion}
    In conclusion, we have presented a novel model for Monocular 3D Object Detection called MonoNext, which uses a ConvNext-based architecture mixed with the MobineNetV2 to develop a lightweight approach for predicting multi-tasks for 3D detections. Our experiments on the KITTI benchmark have shown that the performance of our model is highly dependent on the number of training images. Nevertheless, our method presents competitive performance against state-of-the-art models when tested on the validation set. Moreover, by increasing the training set size, our model outperforms the current models. Although we have only evaluated our model's performance in vehicle detection, it can be easily implemented for different objects in diverse environments. Additionally, the MonoNext model is flexible and capable of detecting multiple objects in the same scene. Overall, this work presents a promising approach for Monocular 3D Object Detection that can be further improved by incorporating additional training data and techniques.

\section{Acknowledge}
The authors acknowledge the funding received from The National Council for Scientific and Technological Development (CNPq 141061/2021-9).

\bibliographystyle{IEEEtran}
\bibliography{root}


\end{document}